\pdfoutput=1

\documentclass[11pt]{article}

\usepackage{emnlp2021}
\usepackage{caption}
\usepackage{times}
\usepackage{latexsym}
\usepackage{graphicx}
\usepackage{subfigure}
\usepackage{amsmath}
\usepackage{multirow}
\usepackage{makecell}
\usepackage{array}
\usepackage{booktabs}
\usepackage{CJK}
\usepackage{hyperref}
\usepackage{float}

\usepackage{amsfonts}
\usepackage{enumitem}

\newcommand{\norm}[1]{\left\lVert #1 \right\rVert}

\usepackage[T1]{fontenc}

\usepackage[utf8]{inputenc}

\usepackage{microtype}

\title{Searching for an Effective Defender: \\ Benchmarking Defense against Adversarial Word Substitution}

 \author{
 Zongyi Li$^{1,2}$\thanks{\ \ Equal contribution}, Jianhan Xu$^{1,2}$\footnotemark[1], Jiehang Zeng$^{1,2}$\footnotemark[1], Linyang Li$^{1,2}$, \\ 
 \textbf{Xiaoqing Zheng$^{1,2}$}, \textbf{Qi Zhang$^{1,2}$}, \textbf{Kai-Wei Chang$^{3}$, Cho-Jui Hsieh$^{3}$}  \\
  $^1$School of Computer Science, Fudan University, Shanghai, China \\
  $^2$Shanghai Key Laboratory of Intelligent Information Processing \\
  $^3$Department of Computer Science, University of California, Los Angeles, USA \\
  
  \texttt{\{zongyili19,jianhanxu20,jhzeng18,zhengxq,qz\}@fudan.edu.cn} \\
  \texttt{\{kwchang,chohsieh\}@cs.ucla.edu}
  }

\begin{document}

\maketitle

\begin{abstract}

Recent studies have shown that deep neural network-based models are vulnerable to intentionally crafted adversarial examples, and various methods have been proposed to defend against adversarial word-substitution attacks for neural NLP models.
However, there is a lack of systematic study on comparing different defense approaches under the same attacking setting. 
In this paper, we seek to fill the gap through comprehensive studies on the behavior of neural text classifiers trained with various defense methods against representative adversarial attacks.
In addition, we propose an effective method to further improve the robustness of neural text classifiers against such attacks, and achieved the highest accuracy on both  clean and adversarial examples on AGNEWS and IMDB datasets, outperforming existing methods by a significant margin. 
We hope this study could provide useful clues for future research on text adversarial defense.
Codes are available at \url{https://github.com/RockyLzy/TextDefender}.

\end{abstract}

\section{Introduction}
Deep neural networks have achieved impressive results on NLP tasks. However, they are vulnerable to intentionally crafted textual adversarial examples, which do not change human understanding of sentences but can easily fool deep neural networks.
As a result, studies on adversarial attacks and defenses in the text domain have drawn significant attention, especially in recent years
\cite{ebrahimi2017hotflip,gao2018black,li2018textbugger,ren2019pwws,jin2020textfooler,li2020bert,jia2019certified,zhu2019freelb,zheng2020evaluating,zhou2020defense,garg-ramakrishnan-2020-bae,zeng2021certified}. 
The goal of adversarial defenses is to learn a model that is capable of achieving high test accuracy on both clean (i.e., original) and adversarial examples.
We are eager to find out which adversarial defense method can improve the robustness of NLP models to the greatest extent while suffering no or little performance drop on the clean input data.



To the best of our knowledge, existing adversarial defense methods for NLP models have yet to be evaluated or compared in a fair and controlled manner. Lack of evaluative and comparative researches impedes understanding of strengths and limitations of different defense methods, thus making it difficult to choose the best defense method for practical use.
There are several reasons why previous studies are not sufficient for comprehensive understanding of adversarial defense methods. Firstly, settings of attack algorithms in previous defense works are far from ``standardized'', and they vary greatly in ways such as synonym-generating methods, number of queries to victim models, maximum percentage of words that can be perturbed, etc.
Most defense methods have only been tested on very few attack algorithms.
Thus, we cannot determine whether one method consistently performs better than others from experimental data reported in the literature, because a single method might demonstrate more robustness to a specific attack while showing much less robustness to another. 
Second, some defense methods work well only when a certain condition is satisfied.
For example, all existing certified defense methods except RanMASK \cite{zeng2021certified} assume that the defenders are informed of how the adversaries generate synonyms \cite{jia2019certified,zhou2020defense,dong2021towards}.
It is not a realistic scenario since we cannot impose a limitation on the synonym set used by the attackers.
Therefore, we want to know which defense method is more effective against existing adversarial attacks when such limitations are removed for fair comparison among different methods.

In this study, we establish a reproducible and reliable benchmark to evaluate the existing textual defense methods, which can provide detailed insights into the effectiveness of defense algorithms with the hope to facilitate future studies.
In particular, we focus on defense methods against adversarial word substitution, one of the most widely studied attack approaches that could cause major threats in adversarial defenses.
In order to rigorously evaluate the performance of defense methods, we propose four evaluation metrics: 
\textit{clean accuracy}, \textit{accuracy under attack}, \textit{attack success rate} and \textit{number of queries}.
The clean accuracy metric measures the generalization ability of NLP models, while the latter three measure the model robustness against adversarial attack.
To systematically evaluate the defense performance of different textual defenders, we first define a comprehensive benchmark of textual attack methods to ensures the generation of high-quality textual adversarial examples, which changes the output of models with human imperceptible perturbation to the input.
We then impose constraints to the defense algorithms to ensure the fairness of comparison. For example, the synonyms set used by adversaries is not allowed to be accessed by any defense method.
Finally, we carry out extensive experiments using typical attack and defense methods for robustness evaluation, including five different attack algorithms and eleven defense methods on both text classification and sentiment analysis tasks.

Through extensive experiments, we found that the gradient-guided adversarial training methods exemplified by PGD \cite{madry2018towards} and FreeLB \cite{zhu2019freelb} can be further improved.
Furthermore a variant of the FreeLB method \cite{zhu2019freelb} outperforms other adversarial defense methods including those proposed years after it.
In FreeLB, gradient-guided perturbations are applied to find the most vulnerable (``worst-case'') points and the models are trained by optimizing loss from these vulnerable points. 
However, magnitudes of these perturbations are constrained by a relatively small constant. 
We find that by extending the search region to a larger $\ell_2$-norm through increasing the number of search steps, much better accuracy can be achieved on both clean and adversarial data in various datasets. 
This improved variant of FreeLB, denoted as FreeLB++, improves the clean accuracy by $0.6\%$ on AGNEWS. FreeLB++ also demonstrates strong robustness under TextFooler attack \cite{jin2020isbert}, achieving a $13.6\%$ accuracy improvement comparing to the current state-of-the-art performance \cite{zeng2021certified}.
Similar results have been confirmed on IMDB dataset.
We believe that our findings invite researchers to reconsider the role of adversarial training, and re-examine the trade-off between accuracy and robustness \cite{zhang2019theoretically}.
Also, we hope to draw attentions on designing adversarial attack and defense algorithms based on fair comparisons.

\section{Background}
\subsection{Textual Adversarial Attacks} \label{adversarial attack}
Textual adversarial attack aims to construct adversarial examples for the purpose of 'fooling' neural network-based NLP models. 
For example, in text classification tasks, a text classifier $f(\boldsymbol{x})$ maps an input text $\boldsymbol{x} \in \mathcal{X}$ to a label $c \in \mathcal{Y}$, where $\boldsymbol{x} = w_1, \dots, w_L$ is a text consisting of $L$ words and $\mathcal{Y}$ is a set of discrete categories. 
Given an original input $\boldsymbol{x}$, an valid adversarial example $\boldsymbol{x}' = w_1',\dots, w_L'$ is crafted to conform to the following requirements:

\begin{equation}
    \small
    f(\boldsymbol{x}') \neq y,  \quad Sim(\boldsymbol{x}, \boldsymbol{x'}) \ge \varepsilon_{min},
\label{eq:adversarial_examples}
\end{equation}
\noindent where $y$ is the ground truth for $\boldsymbol{x}$, $Sim: \mathcal{X} \times \mathcal{X} \rightarrow [0, 1]$ is a similarity function between the original $\boldsymbol{x}$ and its adversarial example $\boldsymbol{x}'$ and $\varepsilon_{min}$ is the minimum similarity. 
In NLP, $Sim$ is often a semantic similarity function using Universal Sentence Encoder (USE) \cite{cer2018universal} to encode two texts into high dimensional vectors and use their cosine similarity score as an approximation of semantic similarity \cite{jin2020textfooler,li2018textbugger}.

\begin{table*}[ht] \small
\setlength{\tabcolsep}{0.5mm}
\setlength{\abovecaptionskip}{0.2cm}
\setlength {\belowcaptionskip} {-0.4cm}
\begin{tabular}{l|p{2.8cm}<{\centering}|l|p{2.0cm}<{\centering}p{1.6cm}<{\centering}p{1.4cm}<{\centering}p{1.5cm}<{\centering}}
\hline
\hline
\multicolumn{3}{c|}{{\bf Method}}  & \makecell{Norm-bounded \\perturbations} & \makecell{Synonyms-\\agnostic} & \makecell{Structure-\\free} & \makecell{Ensemble-\\based}  \\\hline
\multirow{7}{*}{\makecell{\\Empirical\\Defense}} & \multirow{2}{*}{\makecell{Adversarial\\Data Augmentation}} & ADA & & & \checkmark & \\
& & MixADA \cite{si2020better} & & & \checkmark & \\
\cline{2-7}
 
& \multirow{4}{*}{\makecell{\makecell{Adversarial \\Training}}} & PGD-K \cite{madry2018towards} & \checkmark & \checkmark & \checkmark & \\
& & FreeLB \cite{zhu2019freelb} & \checkmark & \checkmark & \checkmark & \\
& & TA-VAT \cite{li2020tavat} & \checkmark & \checkmark & \checkmark & \\
& & InfoBERT \cite{wang2020infobert} & \checkmark & \checkmark & \checkmark & \\
\cline{2-7}
& \multirow{2}{*}{\makecell{\makecell{Region-based\\ Adversarial Training}}} & DNE \cite{zhou2020defense} &  &  & \checkmark & \checkmark \\
& & ASCC \cite{dong2021towards} &  &  & \checkmark &  \\ 
\hline
 
\multirow{3}{*}{\makecell{ \\Certified\\Defense}} & \multirow{2}{*}{\makecell{Interval Bound \\Propagation}} & LSTM-based \cite{jia2019certified} & & &\\
& & Transformer-based \cite{shi2020robustness}  &  &  &  & \\
 \cline{2-7}
 & \multirow{2}{*}{\makecell{Randomized\\Smoothing}} & SAFER\cite{yeetal2020safer} & & & \checkmark & \checkmark \\
& & RanMASK \cite{zeng2021certified}  &  & \checkmark  & \checkmark  & \checkmark \\
\hline
\hline

\end{tabular}

\caption{The comparison of different defense algorithms. We use ``norm-bounded perturbations'' to denote whether the perturbations to word embeddings are norm-bounded, ``synonyms-agnostic'' to whether the defense algorithms rely on pre-defined synonym sets, ``structure-free'' to whether the defense methods can only be applied to specific network architecture, and ``ensemble-based'' to whether the ensemble method is required to produce results.
}
\label{defender_diff}
\end{table*}

\subsection{Adversarial Word Substitution}
Adversarial word substitution is one of the most widely used textual attack methods, where an adversary arbitrarily replaces the words in the original text $\boldsymbol{x}$ by their synonyms according to a synonym set to alert the prediction of the model. 
Specially, for each word $w$, $w' \in S_w$ is any of $w$'s synonyms (including $w$ itself), where the synonym sets $S_w$ are chosen by the adversaries, e.g., built on well-trained word embeddings \cite{mikolov2013efficient,pennington2014glove,su2018exploring}.

The process of adversarial word substitution usually involves two steps: determine an important position to change; and modify words in the selected positions to maximize prediction error of the model. 
To find a word $w' \in  S_w$ that maximizes the model's prediction error, two kinds of searching strategies are introduced: greedy algorithms \cite{kuleshov2018adversarial,li2018textbugger, ren2019generating,hsieh2019robustness,jin2020isbert, li2020bert,yang2020greedy} and combinatorial optimization algorithms \cite{alzantot2018generating, zang2019word}. 
Although the latter usually can fool a model more successfully, they are time-consuming and require massive amount of queries. 
This is especially unfair to defenders, because almost no model can guarantee high prediction accuracy in the case of large-scale queries. 
Therefore, we must impose constraints on the attack algorithm before we systematically evaluate the performance of the defense algorithms, which will be discussed in Section \ref{sec:constraints}.

\subsection{Textual Adversarial Defenses}

Many defense methods have been proposed to improve the robustness of models against text adversarial attacks. Most of these methods focus on defending against adversarial word substitution attack \cite{yeetal2020safer}. 
According to whether they possess provably guaranteed adversarial robustness, these methods can roughly be divided into two categories: \textit{empirical} \cite{zhu2019freelb,zhou2020defense,si2020better,li2020tavat, wang2020infobert,dong2021towards} and \textit{certified defense} \cite{jia2019certified,yeetal2020safer,zeng2021certified} methods. 
Table \ref{defender_diff} demonstrates detailed categories of these defense methods.

\textbf{Adversarial Data Augmentation} (ADA) is one of the most effective empirical defenses \cite{ren2019pwws, jin2020textfooler, li2020bert} for NLP models. 
However, ADA is extremely insufficient due to the enormous perturbation search space, which scales exponentially with the length of input text. 
To cover much larger proportion of the perturbation search space, \citet{si2020better} proposed MixADA, a mixup-based \cite{zhang2017mixup} augmentation method.
\textbf{Region-based adversarial training} \cite{zhou2020defense, dong2021towards} improves a models' robustness by optimizing its performance within the convex hull (Region) formed by embeddings of a word and its synonyms.
\textbf{Adversarial training} \cite{madry2018towards,zhu2019freelb,li2020tavat, wang2020infobert} incorporates a min-max optimization between adversarial perturbations and the models by adding norm-bounded perturbations to words embeddings. 
Previous research on norm-bounded adversarial training focused on improving the generalization of NLP models. 
However, our experimental results showed that these methods can also effectively improve models' robustness while suffering no performance drop on the clean inputs.

It has been experimentally shown that the above empirical methods can defend against attack algorithms. However, they can not provably guarantee that their predictions are always correct even under more sophisticated attackers.
Recently, a set of certified defense methods has been introduced for NLP models, which can be divided into two categories:  \textbf{Interval Bound Propagation} (IBP) \cite{jia2019certified, huang2019achieving, shi2020robustness,xu2020automatic} and \textbf{randomized smoothing} \cite{yeetal2020safer,zeng2021certified} methods. 
IBP-based methods depend on the knowledge of model structure because they compute the range of the model output by propagating the interval constraints of the inputs layer by layer. 
Randomized smoothing-based methods, on the other hand, are structure-free; they constructs stochastic ensembles to input texts and leverage the statistical properties of the ensemble to provably certify the robustness. 
All certified defense methods except RanMASK \cite{zeng2021certified} are based on an assumption that the defender can access the synonyms set used by the attacker. 
Experimental results show that under the same settings, e.g., without accessing the synonyms set, RanMASK achieves the best defense performance among these certified defenders.

\section{Constraints on Adversarial Example Generation}
\label{sec:constraints}



In this section, we first introduce constraints of textual adversarial attacks that should be imposed to ensure the quality of adversarial examples generated, which can help us benchmark textual defense.
Then we introduce the datasets for experiments and pick out the optimal hyper-parameters for each constraint.

\subsection{The Constraints on Adversaries}
To ensure the quality of adversarial examples generated, we impose constraints on textual attack algorithms in the following four aspects:
\begin{itemize}[leftmargin=*]
\setlength{\itemsep}{0pt}
\setlength{\parsep}{0pt}
\setlength{\parskip}{0pt}
\item The minimum semantic similarity $\varepsilon_{min}$ between original input $\boldsymbol{x}$ and adversarial example $\boldsymbol{x}'$.
\item The maximum number of one word's synonyms $K_{max}$.
\item The maximum percentage of modified words  $\rho_{max}$.
\item The maximum number of queries to the victim model $Q_{max}$.
\end{itemize}


\noindent \textbf{Semantic Similarity} 
In order for the generated adversarial examples to be undetectable by human, we need to ensure that the perturbed sentence is semantically consistent with the original sentence.
This is usually achieved by imposing a semantic similarity constraint, see Eq. \eqref{eq:adversarial_examples}. 
Most adversarial attack methods \cite{li2018textbugger, jin2020isbert,li2020bert} use Universal Sentence Encoder (USE) \cite{cer2018universal} to evaluate semantic similarity. USE first encodes sentences into vectors and then uses cosine similarity score between vectors as an approximation of the semantic similarity between the corresponding sentences.
Following the setting in \citet{jin2020textfooler}, we set the default value of minimum semantic similarity $\varepsilon_{min}$ to be $0.84$ \cite{morris2020textattack}.

\begin{figure*} [ht] 
    \centering
    \subfigure[]{
    \label{fig:ag_query}
    \begin{minipage}[t] {0.23\linewidth}
    \includegraphics[width=1.0\linewidth]{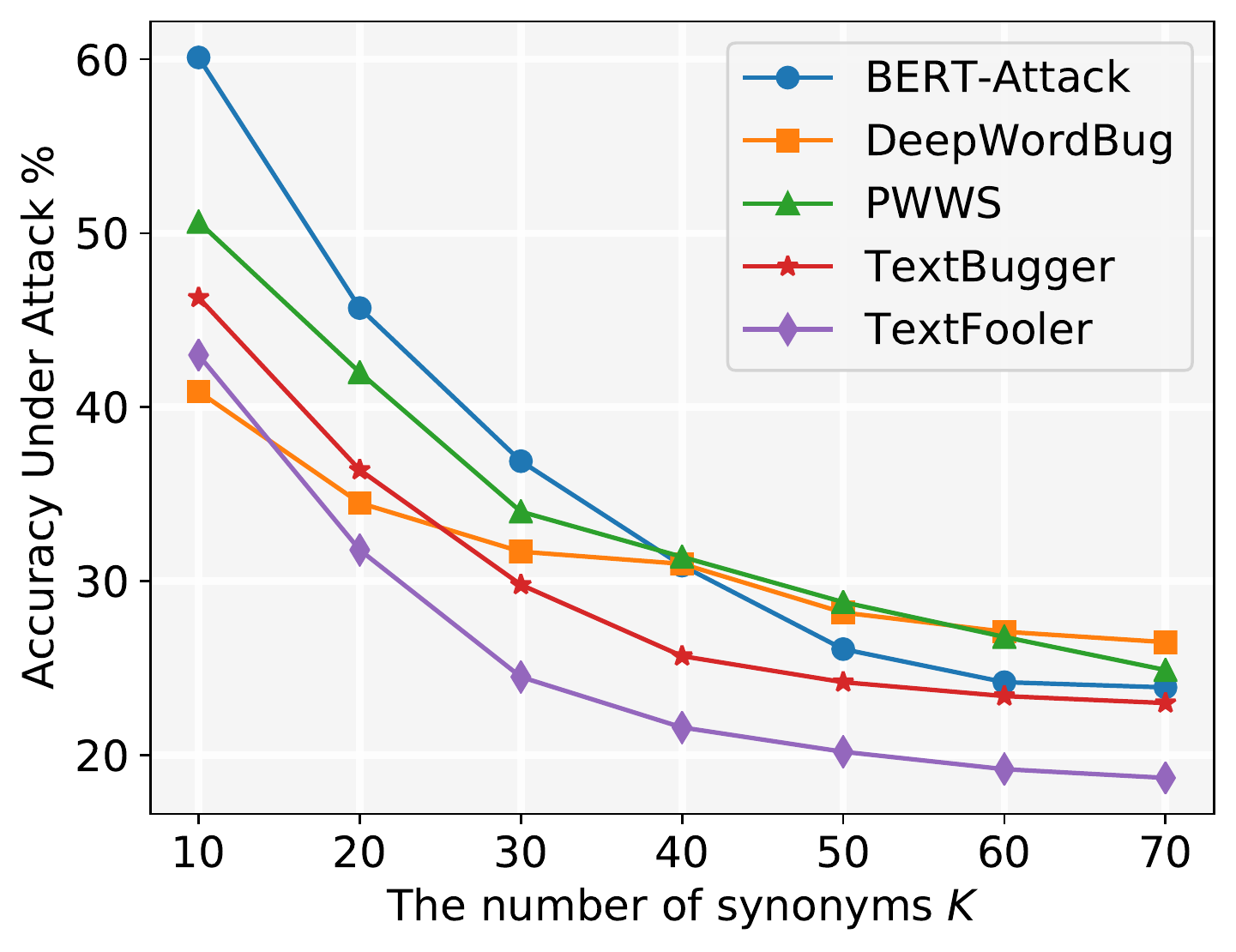}
    \end{minipage}
    }
    \subfigure[]{
    \label{fig:ag_rho}
    \begin{minipage}[t] {0.23\linewidth}
    \includegraphics[width=1.0\linewidth]{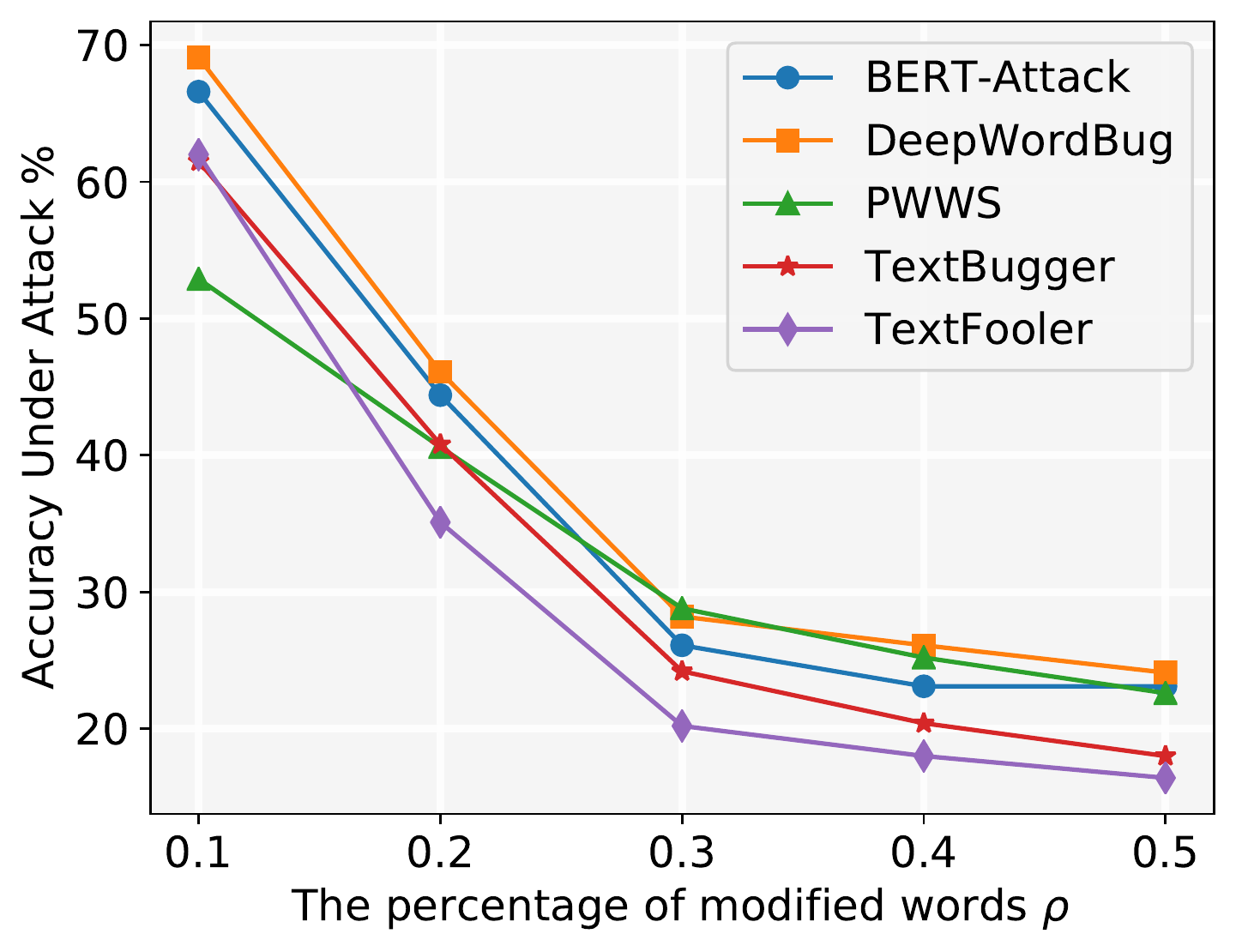}
    \end{minipage}
    }
    \subfigure[]{
    \label{imdb_query}
    \begin{minipage}[t] {0.23\linewidth}
    \includegraphics[width=1.0\linewidth]{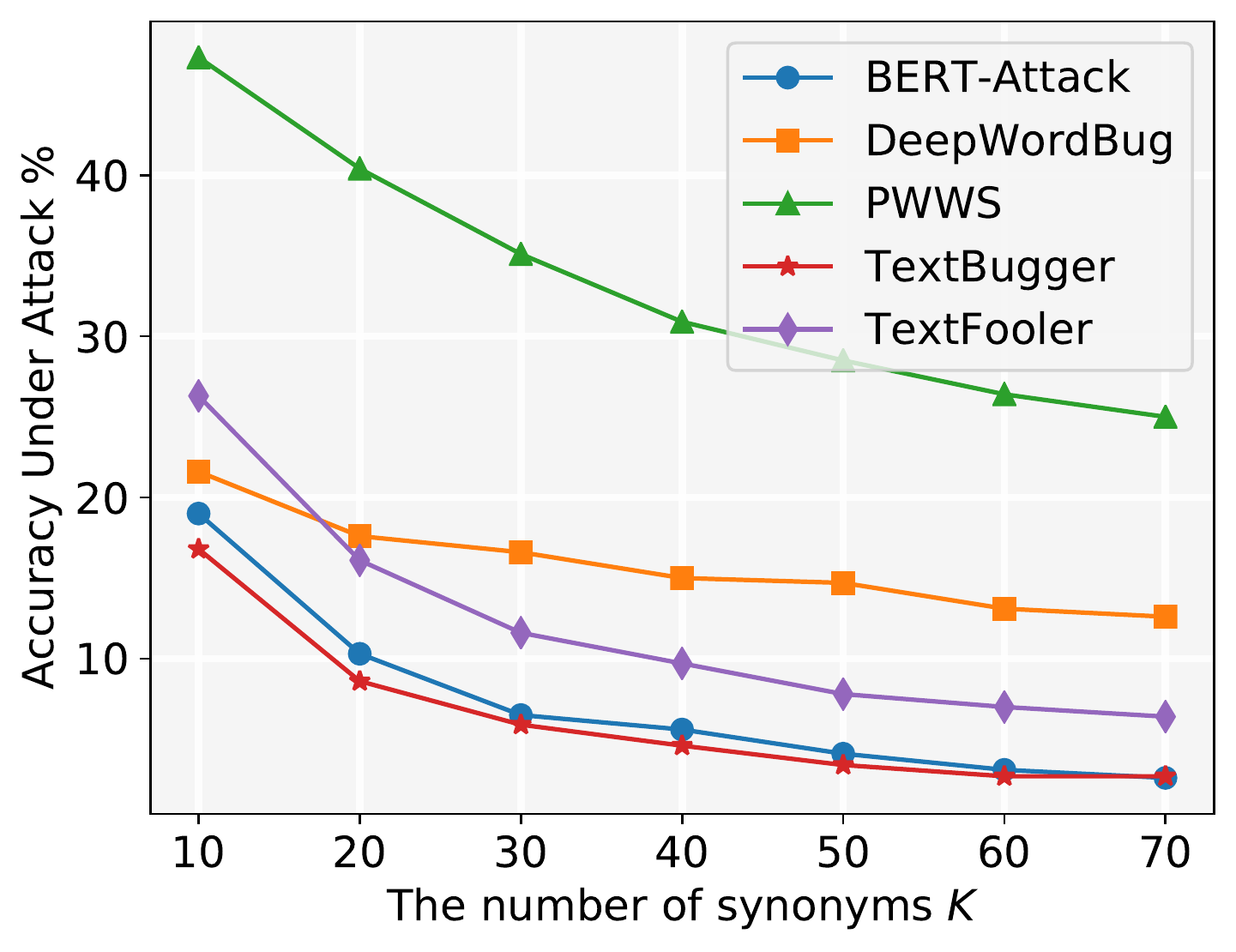}
    \end{minipage}
    }
    \subfigure[]{
    \label{imdb_rho}
    \begin{minipage}[t] {0.23\linewidth}
    \includegraphics[width=1.0\linewidth]{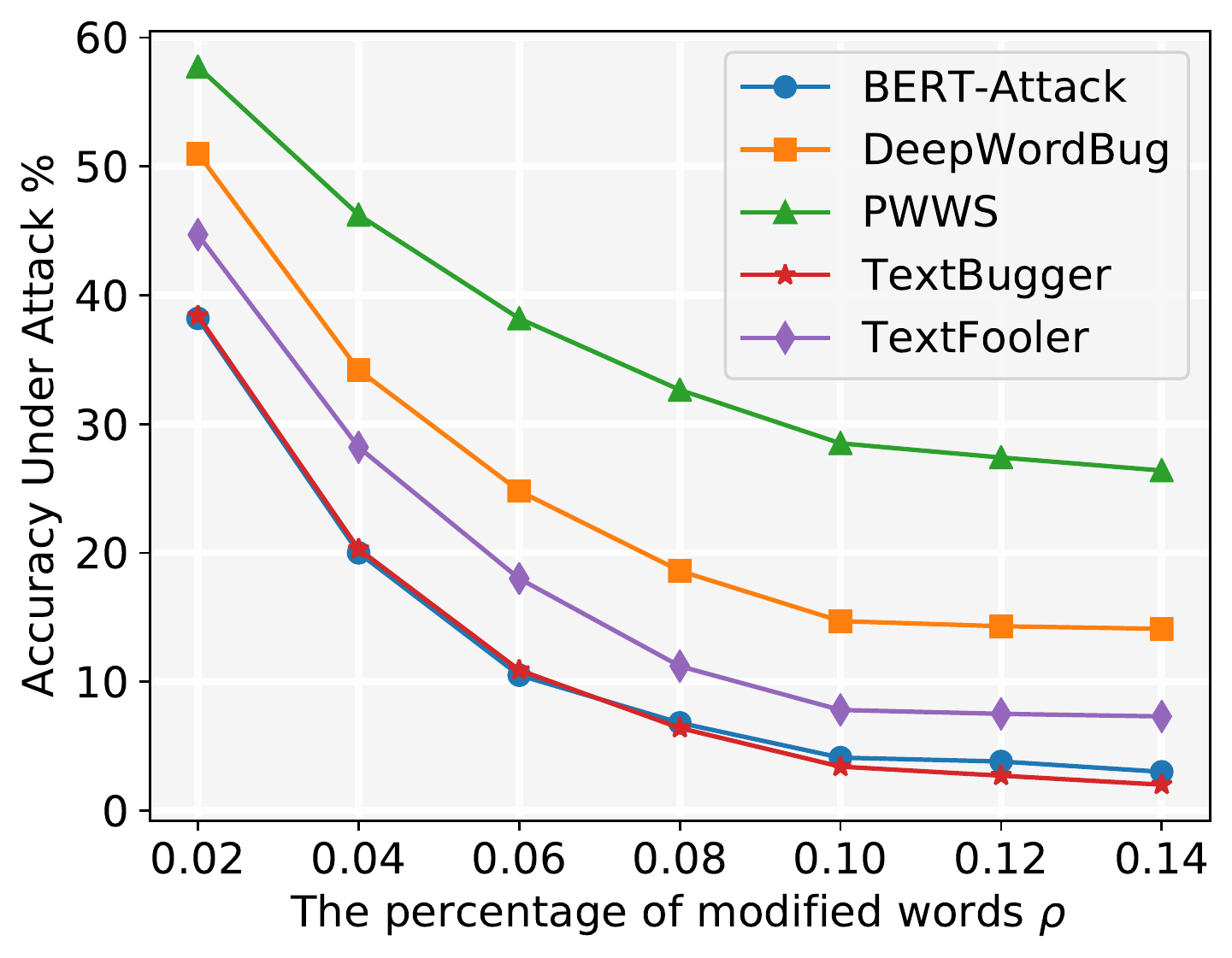}
    \end{minipage}
    }
    \caption{\small The accuracy under attack of five representative greedy search-based attack algorithms with different settings of constraints. Sub-figures (a) and (b) show the impacts of two constraints (the size of synonyms set $K$ and percentage of modified words $\rho$) on AGNEWS, while sub-figures (c) and (d) show the impacts of the same two constraints on IMDB.
    }
    \label{fig:setting}
\end{figure*}

\noindent \textbf{Size of Synonym Set}
For a word $w$ and its synonym set $S_w$, we denote the size of elements in $S_w$ as $K = |S_w|$. 
The value of $K$ influences the search space of attack methods. A larger $K$ increases success rate of the attacker~\cite{morris2020reevaluating}. 
However, larger $K$ would result in the generation of lower-quality adversarial examples since there is no guarantee that these $K$ words are all synonyms of the same word, especially when the GloVe vectors are used to construct a word's synonyms set \cite{jin2020textfooler}. 
While setting the maximum value of $K$ in attack algorithms, we control other variables and select the optimal value $K_{max}$ that keeps attack success rate of the attack algorithm from decreasing too much,
seeing Section \ref{sec:attack_hyper} for more details.

\noindent \textbf{Percentage of Modified Words} 
For an input text $\boldsymbol{x} = w_1, \dots, w_L$, whose length is $L$, and its adversarial examples $\boldsymbol{x}' = w_1', \dots, w_L'$, the percentage of modified words is defined as:
\begin{equation}
    \small
    \rho = \frac{\sum_{i=1}^L{\mathbb{I}\{w_i \neq w_i'\}}}{L},
\label{eq:modified_ratio}
\end{equation}
\noindent where $\sum_{i=1}^L{\mathbb{I}\{x_i \neq x_i'\}}$ is the Hamming distance, with $\mathbb{I}\{ \cdot \}$ being the indicator function. 
An attacker is not allowed to perturb too many words since texcessive perturbation of words results in lower similarity between perturbed and original sentences. 
However, most existing attack algorithms do not limit the modification ratio $\rho$, and sometimes even perturb all words in a sentence to ensure the attack success rate.
Since it is too difficult for defense algorithms to resist such attacks, we restrain a maximum value of $\rho$. 
Similar to the method adopted when setting $K_{max}$, we use control variable to select the optimal value $\rho_{max}$, which will be discussed in Section \ref{sec:attack_hyper}.


\noindent \textbf{Number of Queries}
Some existing attack algorithms achieve high attack success rate through massive queries to the model \cite{yoo2020searching}. 
In order to build a practical attack, we placed constraint on query efficiency.
Considering the difficulty of defense and the time cost of benchmarking, we need to restrict the number of queries for attackers to query the victim model.
At present, most representative attack algorithms are based on greedy search strategies (see Section \ref{adversarial attack}). Experiments have shown that these greedy algorithms are sufficient to achieve a high attack success rate \cite{yang2020greedy}. 
For a greedy-based attack algorithm, assuming the size of its synonyms set is $K = |S_w|$. Then its search complexity is $O(K \times L)$, where $L$ is the length of input text $\boldsymbol{x}$, since the greedy algorithm guarantees that each word in the sentence is replaced at most once.
Thus, we set the maximum number of queries to the product of $K_{max}$ and sentence length $L$ in default, $Q_{max} = K_{max} \times L$. 


\subsection{Datasets and Hyper-parameters}
\label{sec:attack_hyper}
\setlength{\abovecaptionskip}{0.cm}
\setlength {\belowcaptionskip} {-0.6cm}

We conducted experiments on two widely used datasets: the AG-News corpus (AGNEWS) \cite{zhang2015character} for text classification task and the Internet Movie Database (IMDB) \cite{maas2011learning} for sentiment analysis task.

In order to pick the optimal value $K_{max}$ and $\rho_{max}$ for each dataset, we choose $5$ representative adversarial word substitution algorithms: PWWS \cite{ren2019pwws}, TextBugger \cite{li2018textbugger}, TextFooler \cite{jin2020textfooler}, DeepWordBug \cite{gao2018black}, BERT-Attack \cite{li2020bert}. All of them are greedy search based attack algorithms 
\footnote{We also conducted experiments on combinatorial optimization attacker: e.g., GA \cite{alzantot2018generating}, PSO \cite{zang2019word}. However, our pre-experiments showed that they were very time-consuming, and their performance were poor under the limit of the maximum number of queries.}. 
All attackers use $K$ nearest neighbor words of GloVe vectors \cite{pennington2014glove} to generate a word's synonyms except DeepWordBug, which performs character-level perturbations by generating $K$ typos for each word; and BERT-Attack, which dynamically generates synonyms by BERT \cite{devlin2018bert}. We use BERT as baseline model, and implementations are based on TextAttack framework \cite{morris2020textattack}.

When selecting the optimal $K_{max}$ value for AGNEWS, 
we first control other variables unchanged, e.g., the maximum percentage of modified words $\rho_{max}=0.3$, and conduct experiments on AGNEWS with different $K$ values.
As we can see from Figure \ref{fig:ag_query}, as $K$ increases, the accuracy under attack decreases. 
The decline of the accuracy under attack is gradually decreasing.
For $K \ge 50$, the decline in accuracy under attack becomes minimal, thus we pick  $K_{max} = 50$. 
Through the same process, we determine the optimal values $\rho_{max} = 0.1$, $K_{max} = 50$ for IMDB dataset as shown in Figure \ref{imdb_query} and \ref{imdb_rho}, and $\rho_{max} = 0.3$ for AGNEWS dataset as shown in Figure \ref{fig:ag_rho}.

In conclusion, we impose four constraints on attack algorithms to better help with evaluation of different textual defenders. We set $\rho_{max} = 0.3$ for AGNEWS and $\rho_{max} = 0.1$ for IMDB. Such setting is reasonable because the average sentence length of IMDB ($208$ words) is much longer than that of AGNEWS ($44$ words). For other constraints, we set $K_{max}=50$, $\varepsilon_{min} = 0.84$, $Q_{max} = K_{max} \times L$. We choose $3$ base attackers to  benchmark the defense performance of textual defenders: TextFooler, BERT-Attack, and TextBugger. 
Our choice of attackers is based on their outstanding attack performances, as shown in Figure \ref{fig:setting}.



\begin{table*}[!ht] \small
\setlength{\tabcolsep}{1.6mm}
\setlength{\abovecaptionskip}{0.2cm}
\setlength {\belowcaptionskip} {-0.4cm}
\begin{center}
\begin{tabular}{l|c|ccc|ccc|ccc}\hline
\hline

\multicolumn{1}{c|}{\multirow{2}{*}{\bf Method}} & \multicolumn{1}{c|}{\multirow{2}{*}{\bf Clean$\%$}} & \multicolumn{3}{c|}{\bf TextFooler} & \multicolumn{3}{c|}{\bf TextBugger} & \multicolumn{3}{c}{\bf BERT-Attack} \\ \cline{3-11}
\multicolumn{1}{c|}{} & \multicolumn{1}{c|}{} & \multicolumn{1}{p{0.65cm}<\centering}{\bf Aua$\%$} & \multicolumn{1}{p{0.65cm}<\centering}{\bf Suc$\%$} & \multicolumn{1}{c|}{\bf \#Query} & \multicolumn{1}{p{0.65cm}<\centering}{\bf Aua$\%$} & \multicolumn{1}{p{0.65cm}<\centering}{\bf Suc$\%$} &  \multicolumn{1}{c|}{\bf \#Query} & \multicolumn{1}{p{0.65cm}<\centering}{\bf Aua$\%$} & \multicolumn{1}{p{0.65cm}<\centering}{\bf Suc$\%$} & \multicolumn{1}{c}{\bf \#Query} \\ \cline{1-11} 

\hline
Baseline (BERT) &$94.5$ &$19.1$ &$79.6$ &$317.4$ &$23.5$ &$75.0$ &$320.6$ &$27.2$ &$71.0$ &$338.8$ \\
\hline
Adversarial Data Augmentation  &$94.4$ &$38.6$ &$58.9$ &$404.6$ &$43.3$ &$53.9$ &$418.3$ &$42.9$ &$54.5$ &$407.0$ \\
MixADA \cite{si2020better}  &$94.3$ &$37.5$ &$60.3$ &$410.7$ &$36.4$ &$61.4$ &$423.5$ &$39.1$ &$58.6$ &$408.4$ \\
\hline
PGD-K \cite{madry2018towards}  &$94.7$ &$24.8$ &$73.9$ &$353.5$ &$26.7$ &$71.9$ &$367.1$ &$39.4$ &$58.5$ &$399.3$ \\
FreeLB \cite{zhu2019freelb}  &$94.7$ &$31.6$ &$66.7$ &$382.1$ &$32.9$ &$65.4$ &$390.6$ &$43.9$ &$53.8$ &$417.1$ \\
TA-VAT \cite{li2020tavat}  &$94.8$ &$31.0$ &$67.3$ &$382.5$ &$34.2$ &$63.9$ &$415.2$ &$45.0$ &$52.5$ &$436.9$ \\
InfoBERT \cite{wang2020infobert}  &$\bf 95.1$ &$31.8$ &$66.5$ &$369.9$ &$36.3$ &$61.8$ &$391.6$ &$42.4$ &$55.3$ &$392.8$ \\
\hline
DNE \cite{zhou2020defense}  &$93.9$ &$28.7$ &$69.8$ &$367.9$ &$28.2$ &$70.3$ &$377.6$ &$42.4$ &$55.5$ &$470.1$ \\
ASCC \cite{dong2021towards}  &$92.3$ &$28.2$ &$69.6$ &$326.5$ &$37.0$ &$60.1$ &$307.4$ &$32.7$ &$64.7$ &$337.1$ \\
\hline
SAFER \cite{yeetal2020safer}  &$94.3$ &$31.8$ &$66.1$ &$350.1$ &$41.2$ &$56.1$ &$398.8$ &$39.3$ &$58.2$ &$373.5$ \\
RanMASK \cite{zeng2021certified}  &$91.7$ &$37.9$ &$58.7$ &$\bf 583.4$ & $45.0$ & $50.9$ & $\bf 626.8$ &$\bf 49.5$ &$\bf 46.1$ &$\bf 661.8$ \\
\hline
FreeLB++  &$\bf 95.1$ &$\bf 51.5$ &$\bf 46.0$ &$419.1$ &$\bf 55.9$ &$\bf 41.4$ &$416.9$ &$41.8$ &$56.2$ &$386.1$ \\
\hline
\hline
\end{tabular}

\caption{\small \label{agnews_experiment} {The experiment results of different defenders on AGNEWS, where all models are trained on BERT. The best performance is marked in \textbf{bold}. FreeLB++ not only achieves best defense performance under both TextBugger and TextFooler, but also improves \textbf{Clean}\%. Although RanMASK has also achieved significant defense performance, it drops a lot in \textbf{Clean}\%.  }}
\end{center}
\end{table*}

\begin{table*}[!ht]\small
\setlength{\tabcolsep}{1.6mm}
\setlength{\abovecaptionskip}{0.2cm}
\setlength {\belowcaptionskip} {-0.4cm}
\begin{center}
\begin{tabular}{l|c|ccc|ccc|ccc}\hline
\hline

\multicolumn{1}{c|}{\multirow{2}{*}{\bf Method}} & \multicolumn{1}{c|}{\multirow{2}{*}{\bf Clean$\%$}} & \multicolumn{3}{c|}{\bf TextFooler} & \multicolumn{3}{c|}{\bf TextBugger} & \multicolumn{3}{c}{\bf BERT-Attack} \\ \cline{3-11}
\multicolumn{1}{c|}{} & \multicolumn{1}{c|}{} & \multicolumn{1}{p{0.65cm}<\centering}{\bf Aua$\%$} & \multicolumn{1}{p{0.65cm}<\centering}{\bf Suc$\%$} & \multicolumn{1}{c|}{\bf \#Query} & \multicolumn{1}{p{0.65cm}<\centering}{\bf Aua$\%$} & \multicolumn{1}{p{0.65cm}<\centering}{\bf Suc$\%$} &  \multicolumn{1}{c|}{\bf \#Query} & \multicolumn{1}{p{0.65cm}<\centering}{\bf Aua$\%$} & \multicolumn{1}{p{0.65cm}<\centering}{\bf Suc$\%$} & \multicolumn{1}{c}{\bf \#Query} \\ \cline{1-11}

\hline
baseline (BERT) &$92.1$ &$10.3$ &$88.8$ &$488.2$ &$5.3$ &$94.3$ &$438.5$ &$5.8$ &$93.7$ &$412.4$ \\
\hline
Adversarial Data Augmentation  &$91.9$ &$19.0$ &$79.5$ &$837.1$ &$16.1$ &$82.6$ &$910.7$ &$7.4$ &$92.0$ &$436.7$ \\
MixADA \cite{si2020better}  &$91.9$ &$19.0$ &$79.6$ &$523.0$ &$11.5$ &$87.6$ &$518.7 $ &$7.6$ &$91.8$ &$417.47$ \\
\hline
PGD-K \cite{madry2018towards}  &$\bf 93.2$ &$26.0$ &$72.3$ &$577.5$ &$18.9$ &$79.9$ &$624.9$ &$21.0$ &$77.6$ &$525.8$ \\
FreeLB \cite{zhu2019freelb}  &$93.0$ &$29.4$ &$68.3$ &$605.0$ &$22.9$ &$75.3$ &$586.8$ &$21.7$ &$76.6$ &$532.4$ \\
TA-VAT \cite{li2020tavat}  &$93.0$ &$28.2$ &$69.7$ &$606.2$ &$22.8$ &$75.5$ &$681.0$ &$19.2$ &$79.4$ &$486.5$ \\
InfoBERT \cite{wang2020infobert}  &$92.0$ &$19.2$ &$79.2$ &$541.8$ &$12.7$ &$86.3$ &$491.1$ &$11.3$ &$87.8$ &$447.9$ \\
\hline
DNE \cite{zhou2020defense}  &$90.4$ &$28.0$ &$68.2$ &$1222.5$ &$26.5$ &$69.5$ &$1488.0$ &$27.0$ &$69.1$ &$1101.0$ \\
ASCC \cite{dong2021towards}  &$87.8$ &$19.4$ &$77.8$ &$646.1$ &$14.1$ &$83.9$ &$542.5$ &$11.0$ &$87.4$ &$463.2$ \\
\hline
SAFER \cite{yeetal2020safer}  &$91.5$ &$39.5$ &$57.8$ &$1701.7$ &$40.0$ &$57.5$ &$2372.2$ &$38.5$ &$58.8$ &$1363.5$ \\
RanMASK \cite{zeng2021certified}  &$92.3$ &$22.0$ &$74.6$ &$\bf 1493.4$ &$18.0$ &$79.2$ &$1794.9$ &$36.0$ &$58.4$ &$\bf 1813.1$ \\
\hline
FreeLB++  &$\bf 93.2$ &$\bf 45.3$ &$\bf 51.0$ &$1025.9$ &$\bf 42.9$ &$\bf 53.6$ &$1094.0$ &$\bf 39.9$ &$\bf 56.9$ &$696.9$ \\
\hline
\hline
\end{tabular}

\caption{\small \label{imdb_experiment} {The experiment results of different defenders on IMDB. FreeLB++ surpasses all existing defense methods by a large margin under all attackers, even though defending against adversarial attacks in IMDB is harder than that in AGNEWS because sentences from IMDB are far longer than those from AGNEWS. }}
\end{center}
\end{table*}

\section{Experiments on Textual Defense} \label{discussion}
\subsection{Evaluation Metrics}
Under the unified setting of the above-mentioned adversarial attacks, we conducted experiments on the current existing defense algorithms on  AGNEWS and IMDB.
We present $4$ metrics to measure the defense performance.

\begin{itemize}[leftmargin=*]
\setlength{\itemsep}{0pt}
\setlength{\parsep}{0pt}
\setlength{\parskip}{0pt}
    \item The \textit{clean accuracy} (\textbf{Clean}\%) is model's classification accuracy on the clean test dataset.
    
    \item \textit{Accuracy under attack} (\textbf{Aua}\%) is the model's prediction accuracy under specific adversarial attack methods.
    
    \item \textit{Attack success rate} (\textbf{Suc}\%) is the number of texts successfully perturbed by an attack algorithm divided by the number of all texts attempted.
    
    \item \textit{Number of Queries} (\#\textbf{Query}) is the average number of times the attacker queries the model. This is another important metric for evaluating robustness of defenders, since the greater the average query number needed for attacker, the more difficult the defense model is to be compromised.
\end{itemize}

A good defense method should have higher clean accuracy, higher accuracy under attack, lower attack success rate, and requires larger number of queries for attack.

\subsection{Implementation Details}
Our reproduction of all defense methods, along with the hyper-parameter settings, are completely based on their original papers, except for the following two constraints: 
(1) For methods which are not synonyms-agnostic, we establish different synonym sets for both attackers and defender.
(2) For methods that are ensemble-based, we use the ``logit-summed'' ensemble method introduced in \cite{cheng2020voting} to make final predictions.
Specifically, we use the counter-fitting vectors \cite{mrkvsic2016counter} to generate the synonym set for attackers, and use vanilla Glove Embedding \cite{pennington2014glove} to generate synonym set for defenders\footnote{According to our statistics, $69.70\%$ of the words in defender's synonym set appears in the attacker synonym set's vocabulary.
Among them, $73\%$ of the synonyms in the defender's synonym set are covered by the attacker's.}.
Following \citet{cheng2020voting,zeng2021certified}, we take the average of logits produced by the base classifier over all randomly perturbed input sentences, whose size is denoted as $C$, as the final prediction. For AGNEWS, we set the value of $C$ to $100$, while for IMDB, the value of $C$ is default $16$. 
In the implementation of FreeLB++, we remove the constraints of norm bounded projection, and set step size as 30 and 10 on AGNews and IMDB datasets respectively.
More details will be introduced in Section \ref{discussion}.
All the hyper-parameter settings are tuned on a randomly chosen development dataset. 

We use BERT \cite{devlin2018bert} as our base model. Clean accuracy (\textbf{Clean}\%) is tested on the whole test dataset, while the latter three metrics, e.g., \textbf{Aua}\% are evaluated on $1000$ randomly chosen samples from the test dataset.

\subsection{Results} \label{results}

As we can see from Table \ref{agnews_experiment},
(1) the ADA-based methods have a small decrease in clean accuracy, but excellent accuracy under attack.
However, comparing with the remaining methods, ADA-based methods need to know specific attacker algorithms to generate adversarial examples before defending.
(2) The adversarial training methods, e.g., FreeLB, achieve higher clean accuracy than the baseline, and their improvement in robustness is also very insignificant \cite{li2020tavat,zeng2021certified}.
Interestingly, once we remove the $l_2$-norm bounded limitation for FreeLB, we find out that defense performance is significantly improved (see FreeLB++ in the tables).
FreeLB++ surpasses all existing defense methods by a large margin under TextFooler and TextBugger attacks.
We will leave more discussions about adversarial training methods in Section \ref{norm}.
(3) The region-based adversarial training methods, e.g., DNE, perform poorly on both clean accuracy and accuracy under attack.
It is mainly because the synonym set used in the attack method is different from that used in DNE, which is further discussed in Section \ref{synonym}.
(4) The certified defense methods achieve high defense performance.
It is worth noting that the average number of queries to the model of these methods is larger. 
We think the improvement of robustness comes from the ensemble method, seeing further discussions in Section \ref{ensemble discussion}.

Results of defense performance on IMDB are reported in Table \ref{imdb_experiment}.
Defense methods share the trends with performance on AGNEWS.
However, the general robustness of models on IMDB is poorer than AGNEWS.
It is probably because the average length of sentences in IMDB ($208$ words) is far longer than that in AGNEWS ($44$ words).
Longer sentences implies a larger search space for attackers, making it more difficult for defenders to defend against attacks.

\section{Discussions}
\subsection{Effectiveness of Adversarial Training } \label{norm}

The objective of standard adversarial training methods, e.g., PGD-K \cite{madry2018towards} and FreeLB \cite{zhu2019freelb} is to minimize the maximum risk for perturbation $\boldsymbol{\delta}$ within a small $\epsilon$-norm ball:
\begin{equation}
    \small
    \min_{\theta} \mathbb{E}{_{(\boldsymbol{x},y) \sim \mathcal{D}}}  \left[\max_{\lVert \boldsymbol{\delta} \rVert \leq \epsilon}  \mathcal{L}(f_{\theta}(\boldsymbol{X} + \boldsymbol{\delta}), y)\right],
\end{equation}
where $\mathcal{D}$ is the data distribution, $\boldsymbol{X}$ is the embedding representations of input sentence $\boldsymbol{x}$, $y$ is the gold label, and $\mathcal{L}$ is the loss function for training neural networks, whose parameters is denoted as $\theta$. 
In order to solve inner maximization, projected gradient descent (PGD) algorithm is applied as descrided in \citet{madry2018towards} and \citet{zhu2019freelb}:
\begin{equation}
    \small
    \boldsymbol{\delta}_{t+1}=\prod  _{\lVert \boldsymbol{\delta} \rVert _F \leq \epsilon} \left(\boldsymbol{\delta}_t + \alpha \frac{g(\boldsymbol{\delta}_t)}{\lVert g(\boldsymbol{\delta}_t) \rVert _F}\right),
    \label{delta_update}
\end{equation}
where $ g(\boldsymbol{\delta}_t) = \nabla_{\boldsymbol{\delta}}\mathcal{L}(f_{\theta}(\boldsymbol{X} + \boldsymbol{\delta}), y)$ is the gradient of the loss with respect to $\boldsymbol{\delta}$, $\prod  _{\lVert \boldsymbol{\delta} \rVert _F \leq \epsilon}$ performs a projection onto the $\epsilon$-Frobenius norm ball, and $t$ is the number of ascent steps to find the ``worst-case'' perturbation $\boldsymbol{\delta}$ with step size $\alpha$. 

\begin{table}[ht]\small
\setlength{\abovecaptionskip}{0.2cm}
\setlength {\belowcaptionskip} {-0.3cm}
\begin{center}
\resizebox{\linewidth}{!}{
\begin{tabular}{l|c|c|cc|cc}\hline
\hline
\multicolumn{1}{c|}{\multirow{2}{*}{\bf Method}} & \multicolumn{1}{c|}{\multirow{2}{*}{\bf Norm $\epsilon$}} & \multicolumn{1}{c|}{\multirow{2}{*}{\bf Clean\%}} & \multicolumn{2}{c|}{\bf TextFooler} & \multicolumn{2}{c}{\bf BERT-Attack} \\ \cline{4-7}
\multicolumn{1}{c|}{} & \multicolumn{1}{c|}{} & \multicolumn{1}{c|}{} & \multicolumn{1}{c}{\bf Aua\%} & \multicolumn{1}{c|}{\bf Suc\%} &  \multicolumn{1}{c}{\bf Aua\%} & \multicolumn{1}{c}{\bf Suc\%}  \\ \cline{1-7} 
\hline
\multirow{4}*{PGD-K} 

&$0.01$  &$94.9$	&$21.8$	&$77.0$	&$31.5$	&$66.8$\\
&$0.1$   &$95.3$	&$43.6$	&$54.3$	&$45.1$	&$52.7$\\
&$1$ &$\bf 95.2$	&$\bf 45.2$	&$ \bf 55.3$	&$\bf 45.3$	&$\bf 52.4$\\
&w/o &$\bf 95.2$	&$\bf 45.2$	&$\bf 55.3$	&$\bf 45.3$	&$\bf 52.4$ \\
\hline
\multirow{4}*{FreeLB} 

&$0.01$  &$95.4$	&$30.5$	&$68.0$	&$43.6$	&$54.3$\\
&$0.1$   &$95.5$	&$36.1$	&$62.2$	&$40.0$	&$58.1$\\
&$1$ &$\textbf{94.9}$	&$\textbf{45.8}$	&$\textbf{51.7}$	&$\textbf{42.5}$	&$\textbf{55.2}$\\
&w/o &$\textbf{94.9}$	&$\textbf{45.8}$	&$\textbf{51.7}$	&$\textbf{42.5}$	&$\textbf{55.2}$\\
\hline
\hline
\end{tabular}}
\caption{\small The impact of different values of norm $\epsilon$ on both clean and defense performance, where ``w/o'' means updating $\boldsymbol{\delta}$ without projection onto a $\epsilon$-Frobenius norm ball.  A large value $\epsilon = 1$ is equivalent to removing the norm-bounded projection, both of which achieve the best \textbf{Clean}\% and \textbf{Aua}\%.}
\label{norm_exp} 
\end{center}
\end{table}

In this section, we first study the influence of the value of the norm $\epsilon$ on the model's robustness performance, which is also discussed by \citet{gowal2020uncovering} in computer vision field.
As can be seen from Table \ref{norm_exp}, we find out that both of the \textbf{Clean}\% and \textbf{Aua}\% increase as $\epsilon$ increases.
Note that the value of $\epsilon$ is usually set to a very small value, e.g., $\epsilon = 0.01$ \cite{zhu2019freelb}. 
A large value $\epsilon$ (e.g., $\epsilon=1$ in Table \ref{norm_exp}) is equivalent to removing the norm-bounded limitation (seeing ``w/o'' in Table \ref{norm_exp}), because when $\epsilon$ is large enough and the step size $\alpha$ is fixed, the magnitude of perturbation that used to update $\boldsymbol{\delta}$ is also fixed, seeing Eq. \eqref{delta_update}. 
In this case, from $\norm{ \frac{g(\boldsymbol{\delta}_t)}{\lVert g(\boldsymbol{\delta}_t) \rVert _F}}\leq 1$ and Eq. \eqref{delta_update}, 
we have:
\begin{equation}
    \small
    \lVert \boldsymbol{\delta}_{t+1} \rVert \leq \lVert \boldsymbol{\delta}_t \rVert + \left\lVert \alpha \frac{g(\boldsymbol{\delta}_t)}{\lVert g(\boldsymbol{\delta}_t) \rVert _F} \right\rVert \leq \lVert \boldsymbol{\delta}_t \rVert + \alpha.
\end{equation}
Thus, with multi-step updating $\boldsymbol{\delta}$, we have:
\begin{equation} \small
\begin{split}
    \lVert \boldsymbol{\delta}_t \rVert \leq \lVert \boldsymbol{\delta}_{t-1} \rVert + \alpha
    \leq \lVert \boldsymbol{\delta}_{t-2} \rVert + 2 {\alpha} \\
    \leq \cdots \leq \lVert \boldsymbol{\delta}_1 \rVert + (t-1) * \alpha  \leq t \alpha,
\end{split}
\end{equation}


\begin{figure*} [htbp]
\setlength{\abovecaptionskip}{-0.1cm}
\setlength {\belowcaptionskip} {-0.3cm}
    \centering
    \subfigure[]{
    \begin{minipage}[t]{0.31\linewidth}
    \label{fig:clean}
    \includegraphics[width=1.0\linewidth]{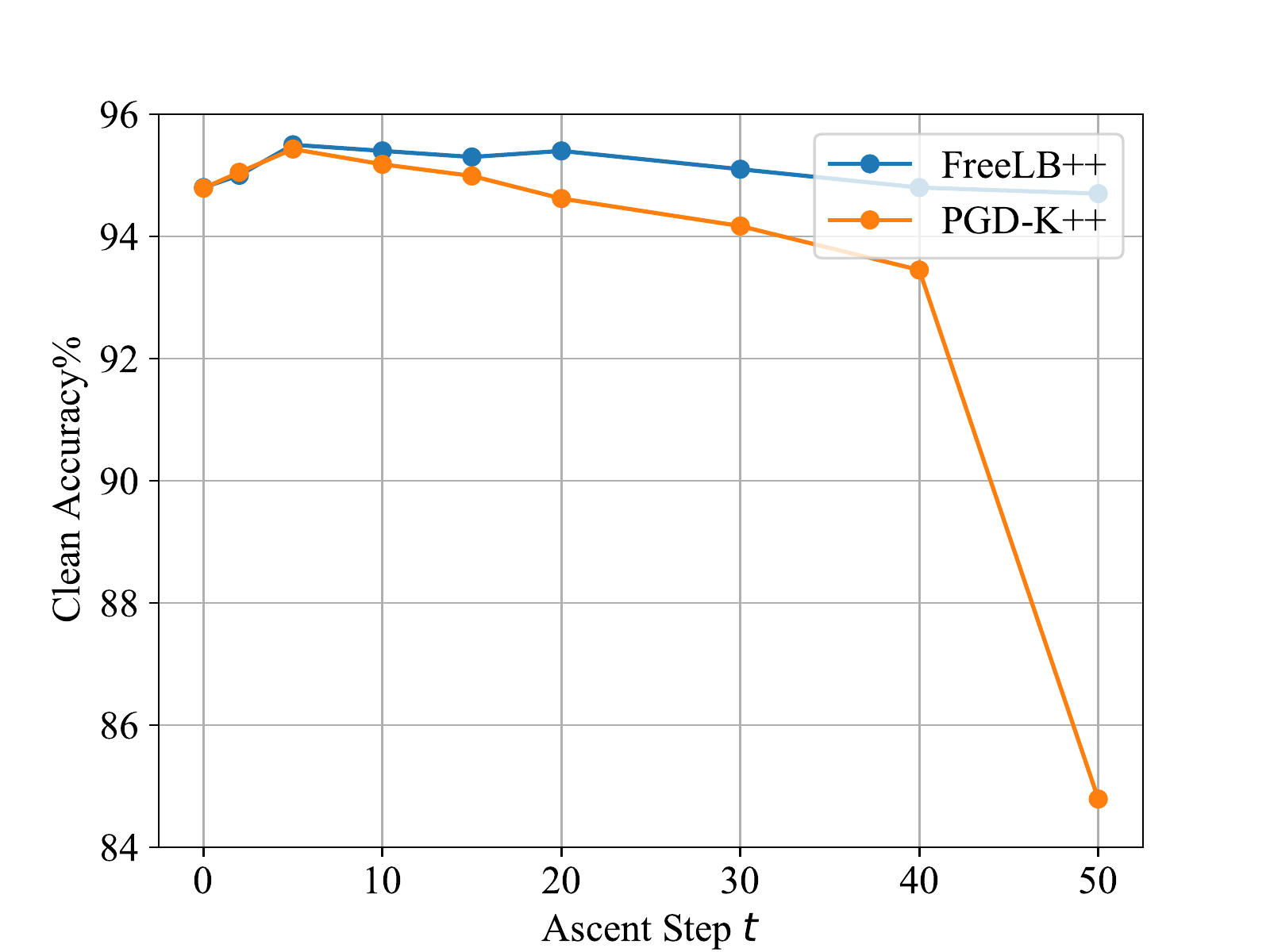}
    \end{minipage}
    }
    \subfigure[]{
    \begin{minipage}[t]{0.31\linewidth}
    \label{fig:steps_freelb}
    \includegraphics[width=1.0\linewidth]{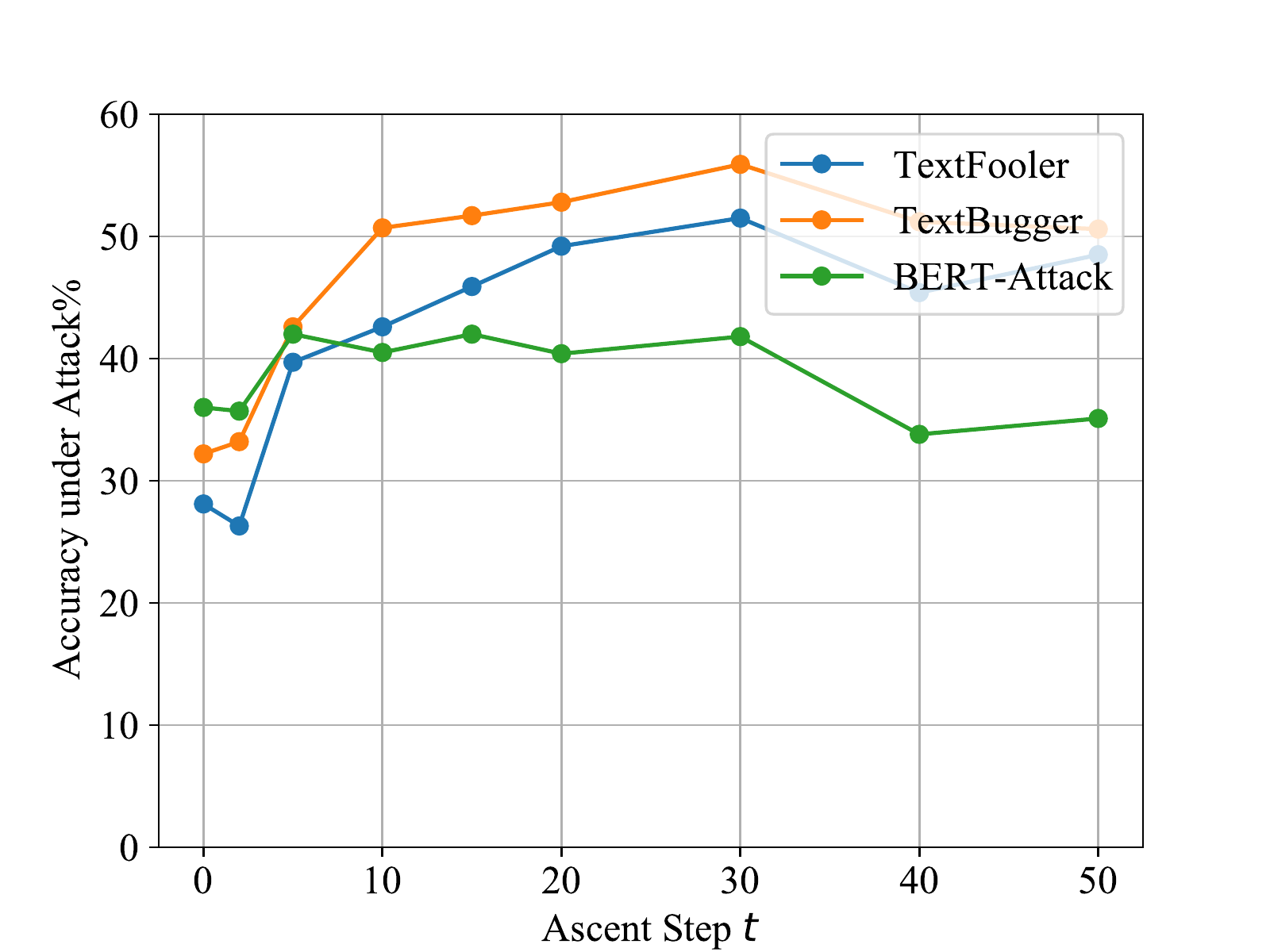}
    \end{minipage}
    }
    \subfigure[]{
    \begin{minipage}[t]{0.31\linewidth}
    \label{fig:steps_pgd}
    \includegraphics[width=1.0\linewidth]{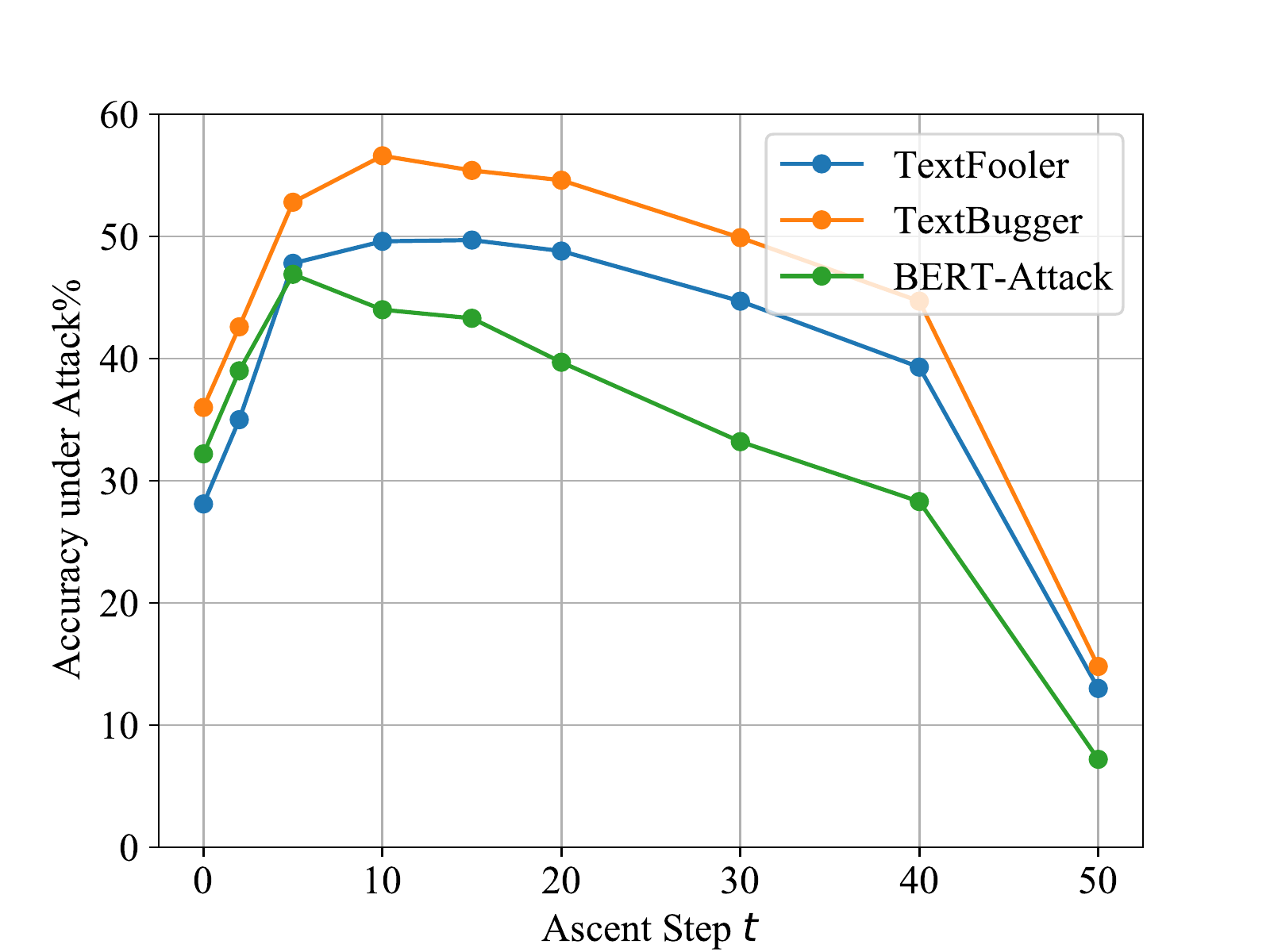}
    \end{minipage}
    }
    \caption{\small The impact of different values of the ascent steps $t$. Sub-figure (a) shows the accuracy of FreeLB++ and PGD-K++ on the clean data (\textbf{Clean\%}). Sub-figure (b) and (c) show the accuracy of FreeLB++ and PGD-K++ respectively under three different attack algorithms (TextFooler, TextBugger and BERT-Attack). As the value of $t$ grows, \textbf{Clean}\% and \textbf{Aua}\% of the models will increase until reaching their peak values. After that, they begin to decrease as the value of $t$ increases continually.
    }
    \label{fig:steps}
\end{figure*}

\noindent where we can find that the upper bound of the norm of perturbation $\boldsymbol{\delta}$ is determined by the number of ascent steps $t$ if the value of $\alpha$ is fixed. In other words, the number of ascent steps $t$ influences the search region of the perturbation $\boldsymbol{\delta}$, where the larger $t$ is, the larger the search region will be. However, in original FreeLB, the same $\epsilon$-norm has been applied to  all perturbations and to restrict the search region around every word embedding. We denote our versions of PGD-K and FreeLB which remove the norm-bounded limitations as PGD-K++ and FreeLB++, respectively. We conducted experiments on PGD-K++, FreeLB++ with different $t$ to study the impact of the value of $t$. 
As shown in Figure \ref{fig:steps}, the \textbf{Clean\%} of both PGD-K++ and FreeLB++ models reaching a peak at $t=5$ while the peak of \textbf{Aua\%} performance reaches at $t=30$ for FreeLB++ and $t=10$ for PGD-K++.

We give a possible explanation to this improvement of performance.
We regard standard adversarial training as an exploration of embedding space.
When $t$ is small, the adversarial example space explored by the model is relatively small, resulting in poor defense performance of the model when a high-intensity attack arrives.
This problem is alleviated when $t$ becomes larger, and this explains why both \textbf{Clean\%} and \textbf{Aua\%} can be improved when $t$ increases.
When $t$ exceeds its optimal value, the adversarial example generated by the algorithm may become dissimilar to the original example.
Excessive learning of examples with different distributions from the original examples will lead to a decline in model's modeling ability.

\subsection{Impact of Ensemble Strategies} \label{ensemble discussion}

There are two ensemble strategies \cite{cheng2020voting}: \textit{logits-summed} (logit) and \textit{majority-vote} (voting) ensemble. 
As mentioned above, in the logit method, the logits produced by the base classifier are averaged. 
Whereas, in the voting strategy, the predictions of classifiers for each class label are counted, and the vote results will be regarded as the output probability for classification. 
Compared to the logit method, we found that the majority-vote strategy can effectively improve the model's robustness, as can be inferred from the results in Table \ref{ensemble_ablation}. 
However, after further research, the reason the voting strategy achieves better defense performance is that it increases the difficulty for score-based attackers, which is also discussed in \citet{zeng2021certified}.

A typical score-based attacker usually involves  two key steps: searching for weak spots in a text and replacing words in these weak spots to maximum model's prediction error.
In the second stage, if no words in the synonym set can lower the logits, the adversary will give up perturbing this word.
However, for those voting-based methods which create ensemble by introducing small noise to the original text $\boldsymbol{x}$, e.g., SAFER, RanMASK-$5\%$, the models tend to output very sharp distribution, even close to one-hot categorical distribution.
This forces the attackers to launch decision-based attacks instead of the score-based
ones, which can dramatically improve their attack difficulty.
Therefore, it may be unfair to compare voting-based ensemble defense methods with others due to lack of effective ways to attack voting-based ensembles in the literature. 
We believe voting-based methods will greatly improve model's defense performance, but we recommend using logit-summed algorithm if one needs to prove the effectiveness of the proposed algorithm against adversarial attacks in future research.

\begin{table}[ht]\small
\begin{center}
\setlength{\abovecaptionskip}{0.2cm}
\setlength {\belowcaptionskip} {-0.5cm}
\resizebox{\linewidth}{!}{
\begin{tabular}{l|c|cc|cc}\hline
\hline
\multicolumn{1}{c|}{\multirow{2}{*}{\bf Method}} &  \multicolumn{1}{c|}{\multirow{2}{*}{\bf Clean}} & \multicolumn{2}{c|}{\bf TextFooler} & \multicolumn{2}{c}{\bf TextBugger} \\ \cline{3-6}
\multicolumn{1}{c|}{} & \multicolumn{1}{c|}{} & \multicolumn{1}{c}{\bf Aua$\%$} & \multicolumn{1}{c|}{\bf Suc$\%$} &  \multicolumn{1}{c}{\bf Aua$\%$} & \multicolumn{1}{c}{\bf Suc$\%$}  \\ \cline{1-6} 

\hline
Baseline (BERT) &$\bf 94.5$ &$19.1$ &$79.6$&$27.2$ &$71.0$ \\
\hline
SAFER (logit) & \multirow{2}{*}{$94.3$}	&$31.8$	&$66.1$	&$41.2$	&$56.1$\\
SAFER (voting) & 	&$ \bf 78.6$	&$ \bf 17.6$	&$\bf 69.0$	&$\bf 28.0$\\
\hline
RanMASK-5\% (logit) & \multirow{2}{*}{$ \bf 94.5$} &$21.5$	&$77.3$	&$41.2$	&$56.1$\\
RanMASK-5\% (voting) & 	& $68.6$ & $26.9$	&$62.5$	&$34.0$\\
\hline
RanMASK-90\% (logit) & \multirow{2}{*}{$91.7$}	&$37.9$	&$58.7$	&$49.5$	&$46.1$\\
RanMASK-90\% (voting) & 	&$47.9$	&$48.2$	&$57.4$	&$37.6$\\
\hline
\hline
\end{tabular}
}
\caption{\small \label{ensemble_ablation} {The ablation experiment on ensemble methods. Voting-based ensembles achieve better performance than logit-based ensembles, but this is potentially due to the non-differentiability introduced by voting-based attacks. 
}}
\end{center}
\end{table}


\subsection{Impact of Synonym Sets} \label{synonym}

Table \ref{vocab} shows the results of the ablation study on the impact of external synonym set on performance in the defense methods.
Some previous studies \cite{yeetal2020safer,zhou2020defense,dong2021towards} use the same synonym set as the attacker during adversarial defense training, leading to significant defense performance.
As we can see from Table \ref{vocab}, all methods improve the \textbf{Aua}\% by a large margin after sharing the synonym set with the attacker.
However, having access to the attacker's synonym is not a realistic scenario since we cannot impose a limitation on the synonym set used by the attackers. 
Thus, for the sake of fair comparison in future research, we suggest that future work should assume that the attacker's synonym set cannot be accessed, and report the defense performance in this case.

\begin{table}[t]\small
\setlength{\abovecaptionskip}{0.1cm}
\setlength {\belowcaptionskip} {-0.5cm}
\begin{center}
\resizebox{\linewidth}{!}{
\begin{tabular}{l|c|cc|cc}\hline
\hline
\multicolumn{1}{c|}{\multirow{2}{*}{\bf Method}} &  \multicolumn{1}{c|}{\multirow{2}{*}{\bf Clean}} & \multicolumn{2}{c|}{\bf TextFooler} & \multicolumn{2}{c}{\bf TextBugger} \\ \cline{3-6}
\multicolumn{1}{c|}{} & \multicolumn{1}{c|}{} & \multicolumn{1}{c}{\bf Aua$\%$} & \multicolumn{1}{c|}{\bf Suc$\%$} &  \multicolumn{1}{c}{\bf Aua$\%$} & \multicolumn{1}{c}{\bf Suc$\%$}  \\ \cline{1-6} 

\hline
SAFER (w) &$93.8$	&$\bf 46.6$	&$\bf 50.3$	&$\bf 55.4$	&$\bf 40.9$ \\
SAFER (w/o) &$\bf 94.3$	&$31.8$	&$66.1$	&$41.2$	&$56.1$\\
\hline
DNE (w) &$93.4$	&$\bf 44.9$	&$\bf 54.0$	&$\bf 42.2$	&$\bf 56.6$\\
DNE (w/o) &$\bf 93.9$	&$28.7$	&$69.8$	&$28.2$	&$70.3$\\
\hline
ASCC (w) &$91.4$	&$\bf 39.0$	&$\bf 57.3$	&$\bf 44.2$	&$\bf 51.6$\\
ASCC (w/o) &$\bf 92.3$	&$28.2$	&$69.6$	&$37.0$	&$60.1$\\
\hline
\hline
\end{tabular}
}
\caption{\small \label{vocab} {The ablation experiment on synonym set. ``w'' and ``w/o'' means the corresponding defense method use or not use the synonym set of the attack method.}}
\end{center}
\end{table}



\section{Conclusion}

In this paper, we established a  comprehensive and coherent benchmark to evaluate the defense performance of textual defenders. 
We impose constraints to existing attack algorithms to ensure the quality of adversarial examples generated. 
Using these attackers, we systematically studied the advantages and disadvantages of different textual defenders. 
We find out that adversarial training methods are still the most effective defenders. 
Our FreeLB++ can not only achieve state-of-the-art defense performance under various attack algorithms, but also improve the performance on clean examples. 
We hope this study could provide useful clues for future research on text adversarial defense.

\section*{Acknowledgements}
This work was supported by 
Shanghai Municipal Science and Technology Major Project (No. 2021SHZDZX0103), and National Science Foundation of China (No. 62076068). CJH is supported by NSF IIS-1901527, IIS-2008173 and IIS-2048280.

\bibliography{anthology,emnlp2021}
\bibliographystyle{acl_natbib}

\end{document}